\documentclass[10pt]{article}
\usepackage[preprint]{tmlr}

\usepackage[utf8]{inputenc} 
\usepackage[T1]{fontenc}    
\usepackage{hyperref}   
\usepackage{url}            
\usepackage{booktabs}       
\usepackage{amsfonts}       
\usepackage{nicefrac}       
\usepackage{microtype}      
\usepackage{xcolor}         
\usepackage{amsmath}
\usepackage{mathtools}
\usepackage{csquotes}
\usepackage{cleveref}
\usepackage{wrapfig}
\usepackage{graphicx}

\title{Convolutional layers are equivariant to discrete shifts\\but not continuous translations}

\author{
\name Nick McGreivy \email mcgreivy@princeton.edu \\
      \addr Department of Astrophysical Sciences\\
  Princeton University\\
  Princeton, NJ
      \AND
      \name Ammar Hakim  \email ahakim@pppl.gov \\
      \addr Princeton Plasma Physics Laboratory \\
   Princeton, NJ
}

\def\month{MM}  
\def\year{YYYY} 
\def\openreview{\url{https://openreview.net/forum?id=XXXX}} 

\begin{document}

\maketitle

\begin{abstract}

The purpose of this short and simple note is to clarify a common misconception about convolutional neural networks (CNNs).
CNNs are made up of convolutional layers which are shift equivariant due to weight sharing.
However, convolutional layers are not translation equivariant, even when boundary effects are ignored and when pooling and subsampling are absent.
This is because shift equivariance is a discrete symmetry while translation equivariance is a continuous symmetry. This fact is well known among researchers in equivariant machine learning, but is usually overlooked among non-experts. To minimize confusion, we suggest using the term `shift equivariance' to refer to discrete shifts in pixels and `translation equivariance' to refer to continuous translations.

\end{abstract}

\section{\label{sec:one} Continuous vs discrete equivariance}

A convolution $\mathcal{C}$ is a linear operator of two functions $a$ and $b$. In one dimension, $\mathcal{C}$ is
\begin{equation}\label{eq:contconvolution}
    \mathcal{C}[a,b](x) = \int_{-\infty}^\infty a(\tau) b(x-\tau) \mathop{d\tau}.
\end{equation}
An operator $f$ is equivariant to a transformation $g$ if \citep{cohen2016group}
\begin{equation}\label{eq:equivariance}
f(g \cdot x) = g \cdot f(x).
\end{equation} 
$\mathcal{C}$ is equivariant to the transformation $x \rightarrow x + \delta$; this is called \textit{translation equivariance}. 

Convolutional layers are the building blocks of convolutional neural networks (CNNs) \citep{zhang1990parallel, lecun1989backpropagation}. Convolutional layers perform a discrete convolution $\mathcal{C}^h$ followed by a nonlinearity $\mathcal{N}^h$ \citep{lecun1995convolutional}. We denote discrete operators and functions with the superscript $h$ and indices with a subscript. A discrete convolution in 1D can be written as
\begin{equation}\label{eq:discreteconvolution}
    \mathcal{C}_j^h[a^h,b^h] = \sum_k a^h_{k} b^h_{j-k}.
\end{equation}
A discrete convolution is equivariant to the transformation $j \rightarrow j + l$; this is called \textit{shift equivariance} \citep{fukushima1982neocognitron,bronstein2021geometric,cohen2016group}. If the nonlinearity $\mathcal{N}^h$ is also shift equivariant, then the convolutional layer $\mathcal{N}^h\big(\mathcal{C}^h[a^h,b^h]\big)$ will be shift equivariant, ignoring boundary effects \citep{azulay2018deep,kayhan2020translation, zhang2019making, chaman2021truly}.

We now show that these layers are not translation equivariant. The essence of the argument is that translation equivariance is a property of continuous systems, while convolutional layers operate on discrete models that do not have a continuous symmetry. 

The data from the real-world system $f(x)$ is a continuous function. To map from the continuous system to the discrete model, we introduce a discretization operator $\mathcal{D}^h$, where $\mathcal{D}^h[f(x)] = f^h$. In general, it is not possible to map from the discrete model back to the continuous system. Applying a convolutional layer to the continuous data $f(x)$ can thus be written as
$\mathcal{N}^h\big(\mathcal{C}^h[a^h, \mathcal{D}^h [f(x)]]\big)$ where $\mathcal{N}^h$ is the nonlinearity and $a^h$ is the convolutional kernel.
By the definition of equivariance in \cref{eq:equivariance}, the convolutional layer is translation equivariant if
\begin{equation}\label{eq:equivariancelayer}
    \mathcal{N}^h\big(\mathcal{C}^h[a^h, \mathcal{D}^h [f(g \cdot x)]]\big) \stackrel{?}{=} g \cdot \mathcal{N}^h\big(\mathcal{C}^h[a^h, \mathcal{D}^h [f(x)]]\big)
\end{equation}
\looseness=-1 where $g$ is the transformation $x \rightarrow x + \delta$ for $\delta \in \mathbb{R}$. The left hand side of \cref{eq:equivariancelayer} is well-defined; it involves translating $f(x)$ by $\delta$, discretizing $f(x+\delta)$, then performing the convolution and nonlinearity. However, the right hand side of \cref{eq:equivariancelayer} is not well-defined; it requires translating a discrete quantity by a continuous amount. Therefore, \cref{eq:equivariancelayer} cannot possibly be true, meaning that convolutional layers are not translation equivariant.

Strictly speaking, it is possible to define a discrete translation $g^h$ which translates discrete data by a non-integer number of pixels. A discrete translation $g^h$ could be defined, for example, by interpolating the discrete data between gridpoints, translating the interpolated data, then discretizing the result.
Nevertheless, it is is impossible to design discrete systems that have an exact continuous translation equivariance, i.e.,
\begin{wrapfigure}[17]{r}{0.4\textwidth}
    \centering
    \vspace{-1\intextsep}
    \setlength{\columnsep}{0pt}
    \setlength\intextsep{0pt}
    \includegraphics[width=0.4\textwidth]{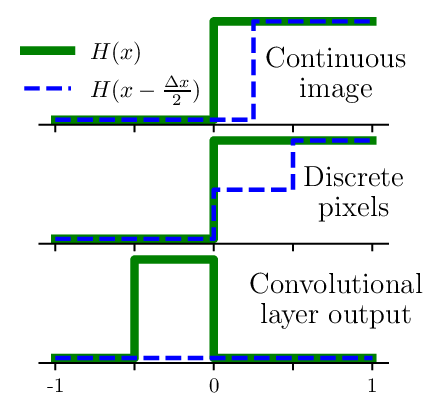}
    \vspace{-1.0cm}
    \caption{While the convolutional layer detects an edge in the original image $H(x)$, it doesn't detect an edge in the translated image $H(x-\nicefrac{\Delta x}{2})$.}
    \label{fig:convolution}
\end{wrapfigure}
\begin{equation}\label{eq:discreteg}
    \mathcal{D}^h[f(g \cdot x)] \ne g^h \cdot \mathcal{D}^h[f(x)],
\end{equation}
because information about the continuous function $f(x)$ is lost in the discretization process. It is possible to design discrete convolutions that do not change significantly (i.e., are Lipschitz continuous) to perturbations of the input \citep{bruna2013invariant}. 

\section{\label{sec:implications}Intuition}

We consider a simple example of an image in 1D. Suppose our image domain is $x \in [-1, 1]$ and our 1D image is the Heaviside step function $H(x)$ where
\begin{equation*}
    H(x) \coloneqq    
    \begin{cases}
      1 & \text{if $x>0$}\\
      0 & \text{if $x \le 0$.}
    \end{cases}   
\end{equation*}

Now suppose we discretize (i.e., `take a picture of') our image $H(x)$ using a discretization operator which computes the average value of the image $H^h_j$ inside the $j$th pixel for $j=0,\dots N-1$.
This means that $H^h_j=\mathcal{D}_j^h[H(x)] = \int_{x_j}^{x_{j+1}} H(x) \mathop{dx}$, where $x_j = -1 + j \Delta x$ are the pixel boundaries and $\Delta x = \frac{2}{N}$. The image has $N=4$ pixels. Suppose also that our convolutional layer performs a convolution with kernel $a_k^h=[2,-2]$ and bias $-1$ followed by a ReLU nonlinearity; this layer is designed to detect edges in the image.

Now, let us compare the output of the convolutional layer between the image $H(x)$ and a translated image $H(x - \frac{\Delta x}{2})$. The original image pixels are $\mathcal{D}^h[H(x)] = [0, 0, 1, 1]$, while the translated image pixels are  $\mathcal{D}^h[H(x-\frac{\Delta x}{2})] = [0, 0, 0.5, 1]$. As illustrated in \cref{fig:convolution}, the output of the convolutional layer on the original image is $[0,1,0,0]$ while the output of the convolutional layer on the translated image is $[0,0,0,0]$. The convolutonal layer detects an edge in the first image, but does not detect an edge in the translated image. This example demonstrates intuitively the point of this paper: that convolutional layers are equivariant to discrete shifts in pixels, but not equivariant to continuous translations in images.


\bibliographystyle{tmlr}
\bibliography{bibliography}

\end{document}